\documentclass[runningheads]{llncs}
\usepackage{graphicx}
\usepackage{multirow}
\usepackage{booktabs}
\usepackage{bbding}
\usepackage{hyperref, color}

\begin{document}
\title{Rethinking Surgical Instrument Segmentation: A Background Image Can Be All You Need}
\titlerunning{A Background Image Can Be All You Need for Instrument Segmentation}
%
\author{An Wang\inst{1,2~\star} 
\and Mobarakol Islam\inst{3} 
\thanks{An Wang and Mobarakol Islam are co-first authors.} 
\and Mengya Xu\inst{4} 
\and Hongliang Ren\inst{1,2,4} 
\thanks{Corresponding author.}}
%
\authorrunning{A. Wang et al.}
%
\institute{Dept. of Electronic Engineering, The Chinese University of Hong Kong, Hong Kong SAR, China
\and Shun Hing Institute of Advanced Engineering, The CUHK, Hong Kong SAR, China
\and BioMedIA Group, Dept. of Computing, Imperial College London, London, UK
\and Dept. of Biomedical Engineering, National University of Singapore, Singapore\\
\email{wa09@link.cuhk.edu.hk, m.islam20@imperial.ac.uk, mengya@u.nus.edu, hlren@ee.cuhk.edu.hk}}
\maketitle              
\begin{abstract}
Data diversity and volume are crucial to the success of training deep learning models, while in the medical imaging field, the difficulty and cost of data collection and annotation are especially huge. Specifically in robotic surgery, data scarcity and imbalance have heavily affected the model accuracy and limited the design and deployment of deep learning-based surgical applications such as surgical instrument segmentation. Considering this, we rethink the surgical instrument segmentation task and propose a one-to-many data generation solution that gets rid of the complicated and expensive process of data collection and annotation from robotic surgery. In our method, we only utilize a single surgical background tissue image and a few open-source instrument images as the seed images and apply multiple augmentations and blending techniques to synthesize amounts of image variations. In addition, we also introduce the chained augmentation mixing during training to further enhance the data diversities. The proposed approach is evaluated on the real datasets of the EndoVis-2018 and EndoVis-2017 surgical scene segmentation. Our empirical analysis suggests that without the high cost of data collection and annotation, we can achieve decent surgical instrument segmentation performance. Moreover, we also observe that our method can deal with novel instrument prediction in the deployment domain. We hope our inspiring results will encourage researchers to emphasize data-centric methods to overcome demanding deep learning limitations besides data shortage, such as class imbalance, domain adaptation, and incremental learning. Our code is available at~\url{https://github.com/lofrienger/Single_SurgicalScene_For_Segmentation}.

\end{abstract}
\section{Introduction}
\label{sec:intro}
Ever-larger models processing larger volumes of data have propelled the extraordinary performance of deep learning-based image segmentation models in recent decades, but obtaining well-annotated and perfectly-sized data, particularly in the medical imaging field, has always been a great challenge~\cite{domingos2012few}. Various causes, including tremendous human efforts, unavailability of rare disease data, patient privacy concerns, high prices, and data shifts between different medical sites, have made acquiring abundant high-quality medical data a costly endeavor. Besides, dataset imperfection like class imbalance, sparse annotations, noisy annotations and incremental-class in deployment~\cite{xu2021class} also affects the training and deployment of deep learning models. Moreover, for the recent-developed surgery procedures like the single-port robotic surgery where no dataset of the new instruments is available~\cite{dobbs2020single}, the segmentation task can hardly be accomplished. In the presence of these barriers, one effective solution to overcome the data scarcity problems is to train with a synthetic dataset instead of a real one. 

A few recent studies utilize synthetic data for training and achieve similar and even superior performance than training with real data. For example, in the computer vision community, Tremblay et al.~\cite{tremblay2018training} develop an object detection system relying on domain randomization where pose, lighting, and object textures are randomized in a non-realistic manner; Gabriel et al.~\cite{eilertsen2021ensembles} make use of multiple generative adversarial networks (GANs) to improve data diversity and avoid severe over-fitting compared with a single GAN; Kishore et al.~\cite{kishore2021synthetic} propose imitation training as a synthetic data generation guideline to introduce more underrepresented items and equalize the data distribution to handle corner instances and tackle long-tail problems.

In medical applications, many works have focused on GAN-based data synthesizing~\cite{shin2018medical,cao2020auto,han2019synthesizing,hamghalam2020high}, while a few works utilize image blending or image composition to generate new samples. For example, mix-blend~\cite{garcia2021image} mixes several synthetic images generated with multiple blending techniques to create new training samples. Nonetheless, one limitation of their work is that they need to manually capture and collect thousands of foreground instrument images and background tissue images, making the data generation process trivial and time-consuming. 
In addition, E. Colleoni et al.~\cite{colleoni2020synthetic} recorded kinematic data as the data source to synthesize a new dataset for the instrument - Large Needle Drivers. 
In comparison with previous works, our approach only utilizes a single background image and dozens of foreground instrument images as the data source. Without costly data collection and annotation, we show the simplicity and efficacy of our dataset generation framework. 

\subsubsection{Contributions}
In this work, we rethink the surgical instrument segmentation task from a data-centric perspective. Our contributions can be summarized as follows:
\begin{itemize}
    \item With minimal human effort in data collection and without manual image annotations, we propose a data-efficient framework to generate high-quality synthetic datasets used for surgical instrument segmentation.
    \item By introducing various augmentation and blending combinations to the foreground and background source images, and training-time chained augmentation mixing, we manage to increase the data diversity and balance the instruments class distribution.
    \item We evaluate our method on two real datasets. The results suggest that our dataset generation framework is simple yet efficient. It is possible to achieve acceptable surgical instrument segmentation performance, even for novel instruments, by training with synthetic data that only employs a single surgical background image.
\end{itemize}

\section{Proposed Method} 
\subsection{Preliminaries}
\textbf{Data augmentation} has become a popular strategy for boosting the size of a training dataset to overcome the data-hungry problem when training the deep learning models. Besides, data augmentation can also be regarded as a regularisation approach for lowering the model generalization error~\cite{goodfellow2016deep}. In other words, it helps boost performance when the model is tested on a distinct unseen dataset during training. Moreover, the class imbalance issue, commonly seen in most surgical datasets, can also be alleviated by generating additional data for the under-represented classes. 

\begin{figure}[!hbpt]
\centering
\includegraphics[width=0.95\textwidth]{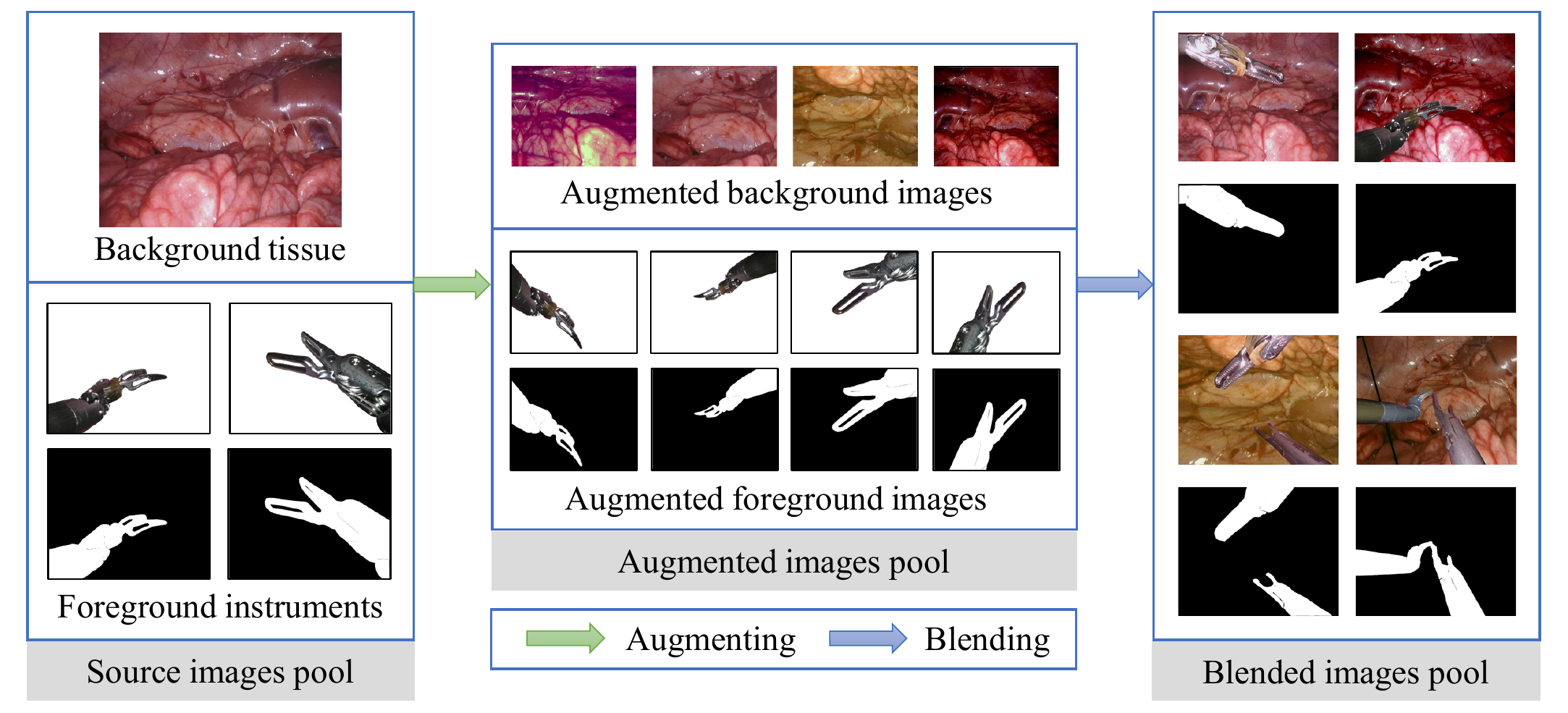}
\caption{\textbf{Demonstration of the proposed dataset generation framework with augmenting and blending.} With minimal effort in preparing the source images, our method can produce large amounts of high-quality training samples for the surgical segmentation task.
}
\centering
\label{fig:blend}
\end{figure}

\textbf{Blending} is a simple yet effective way to create new images simply by image mixing or image composition. It can also be treated as another kind of data augmentation technique that mixes the information contained in different images instead of introducing invariance to one single image. Denote the foreground image and background image as $x_f$ and $x_b$, we can express the blended image with a blending function $\Theta$ as 
\begin{equation}
    x=\Theta(x_f, x_b)=x_f \oplus x_b
    \label{eq:blend}
\end{equation}
where $\oplus$ stands for pixel-wise fusion.

\textbf{Training-time augmentation} can help diversify training samples. By mixing various chained augmentations with the original image, more image variations can be created without deviating too far from the original image, as proposed by AugMix~\cite{hendrycks2019augmix}. In addition, intentionally controlling the choices of augmentation operations can also avoid hurting the model due to extremely heavy augmentations. 
A list of augmentation operations is included in the augmentation chains, such as auto-contrast, equalization, posterization, solarization, etc.

\subsection{Synthesizing surgical scenes from a single background}

\subsubsection{Background tissue image processing}
We collect one background tissue image from the open-source EndoVis-2018 dataset~\footnote{\url{https://endovissub2018-roboticscenesegmentation.grand-challenge.org/}} where the surgical scene is the nephrectomy procedures. The critical criterion of this surgical background selection is that the appearance of the instrument should be kept as little as possible. In the binary instrument segmentation task, the background pixels are all assigned with the value 0. Therefore, the appearance of instruments in the source background image will occupy additional effort to handle. Various augmentations have been applied to this single background source image with the imgaug~\footnote{\url{https://github.com/aleju/imgaug}} library~\cite{imgaug}, including LinearContrast, FrequencyNoiseAlpha, AddToHueAndSaturation, Multiply, PerspectiveTransform, Cutout, Affine, Flip, Sharpen, Emboss, SimplexNoiseAlpha, AdditiveGaussianNoise, CoarseDropout, GaussianBlur, MedianBlur, etc. We denote the generated $p$ variations of the background image as the background images pool $X_b^p=\{x_b^1, x_b^2,..., x_b^p\}$. As shown in Fig.~\ref{fig:blend}, various augmented background images are generated from the single source background tissue image to cover a wide range of background distribution.

\subsubsection{Foreground instruments images processing}
We utilize the publicly available EndoVis-2018~\cite{allan20202018} dataset as the open resource to collect the seed foreground images. There are 8 types of instruments in the EndoVis-2018~\cite{allan20202018} dataset, namely Maryland Bipolar Forceps, Fenestrated Bipolar instruments, Prograsp Forceps, Large Needle Driver, Monopolar Curved Scissors, Ultrasound Probe, Clip Applier, and Suction Instrument. We only employ 2 or 3 images for each instrument as the source images. We extract the instruments and make their background transparent. The source images are selected with prior human knowledge of the target scenes to ensure their high quality. For example, for some instruments like Monopolar Curved Scissors, the tip states (open or close) are crucial in recognition, and they are not reproducible simply by data augmentation. Therefore, we intentionally select source images for such instruments to make it possible to cover different postures and states. In this way, we aim to increase the in-distribution data diversity to substantially improve generalization to out-of-distribution (OOD) category-viewpoint combinations~\cite{madan2020and}. Since we get rid of annotation, the instrument masks are applied with the same augmentations as the instruments to maintain the segmentation accuracy. We denote the generated $q$ variations of the foreground images as the foreground image pool $X_f^q=\{x_f^1, x_f^2,..., x_f^q\}$.
Figure~\ref{fig:blend} shows some new synthetic instruments images. The foreground images pool, together with the background images pool, forms the augmented images pool, which is used for the following blending process. 

\subsubsection{Blending images}
After obtaining the background image pool $X_b^p$ and the foreground image pool $X_f^q$, we randomly draw one sample from these two pools and blend them to form a new composited image. Specifically, the foreground image is pasted on the background image with pixel values at the overlapped position taken from the instruments. Furthermore, considering the real surgical scenes, the number of instruments in each image is not fixed. We also paste two instrument images on the background occasionally. Due to this design, we expect the model could better estimate the pixel occupation of the instruments in the whole image. Denoting the blended image as $x_s$, finally, the blended images pool with $t$ synthetic images can be presented as $X_s^t=\{x_s^1, x_s^2,..., x_s^t\}=\{\Theta(x_f^i, x_b^j)\}$, where $i=1,2,...,p$ and $j=1,2,...,q.$

\subsubsection{In-training chained augmentation mixing}
Inspired by AugMix~\cite{hendrycks2019augmix}, we apply the training-time chained augmentation mixing technique to further make the data more diverse and also improve the generalization and robustness of the model. The number of augmentation operations in each augmentation chain is randomly set as one, two, or three. The parameters in the Beta distribution and the Dirichlet distribution are all set as 1. We create two sets of augmentation collections, namely AugMix-Soft and AugMix-Hard. Specifically, AugMix-Soft includes autocontrast, equalize, posterize and solarize, while AugMix-Hard has additional color, contrast, brightness, and sharpness augmentations. 
The overall expression of the synthetic training sample after the training-time augmentation mixing with $N$ chains is 
\begin{equation}
    x_s^{AM} = m \cdot \Theta(x_f, x_b) + (1-m) \cdot \sum_{i=1}^{N}{(w_i\cdot H_i(\Theta(x_f, x_b)))}
    \label{eq:augmixsyn}
\end{equation}
where $m$ is a random convex coefficient sampled from a Beta distribution, $w_i$ is also a random convex coefficient sampled from a Dirichlet distribution controlling the mixing weights of the augmentation chains. Both distribution functions have the same coefficient value of 1. $H_i$ denotes the integrated augmentation operations in the $i^{th}$ augmentation chain. 

\section{Experiments}
\subsection{Datasets}
Based on effortlessly collected source images and considering the contents in real surgery images, we apply a wide range of augmentation and blending operations to create abundant synthetic images for training. Only one background tissue image is adopted to generate our synthetic datasets. Specifically, for the case of 2 source images per instrument, we first organize the dataset Synthetic-A with 4000 synthetic images, and only one instrument exists in each synthetic image. Then we consider adding up additional 2000 synthetic images to build the dataset Synthetic-B where each image contains 2 distinct instruments. Moreover, we utilize one more source foreground image for each instrument and generate 2000 more synthetic images, among which 80\% contain one instrument, and the remaining 20\% contain 2 different instruments. This dataset with 8000 samples in total is named Synthetic-C. 

To evaluate the quality of the generated surgical scene dataset, we train a binary segmentation model with our synthetic datasets and the training set of the EndoVis-2018~\cite{allan20202018} dataset. The model is tested with the test set of this real dataset. We also test with the whole training set of EndoVis-2017~\cite{allan20192017} dataset to show that the model trained with our synthetic dataset also obtains good generalization ability to handle new domains with unseen instruments like the Vessel Sealer.

\begin{figure}[!hbpt]
\centering
\includegraphics[width=0.8\textwidth]{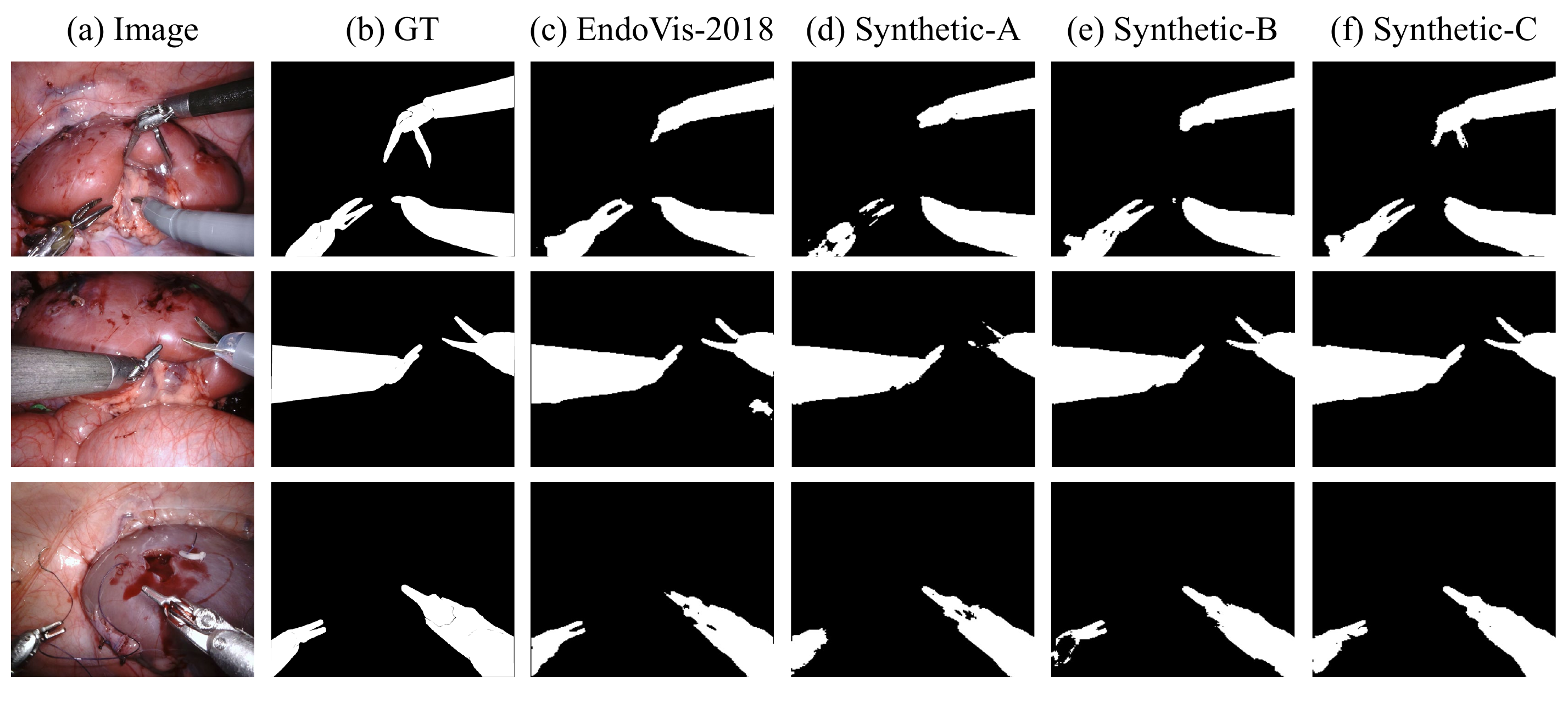}
\caption{\textbf{Qualitative comparison of the binary segmentation results.} (b) represents the ground truth. (c), (d), (e), and (f) show the results obtained from models trained with the EndoVis-2018 dataset and our three synthetic datasets. 
}
\centering
\label{fig:qual_comp}
\end{figure}

\subsection{Implementation details}
The classic state-of-the-art encoder-decoder network UNet~\cite{ronneberger2015unet} is used as our segmentation model backbone. We adopt a vanilla UNet architecture~\footnote{\url{https://github.com/ternaus/robot-surgery-segmentation}} with Pytorch~\cite{paszke2017automatic} library and train the model with NVIDIA RTX3090 GPU. The batch size of 64, the learning rate of 0.001, and the Adam optimize are identically used for all experiments. The binary cross-entropy loss is adopted as the loss function. We use the Dice Similarity Coefficient (DSC) to evaluate the segmentation performance. The images are resized to 224$\times$224 to save the training time. Besides, we refer to the implementation~\footnote{\url{https://github.com/google-research/augmix}} of AugMix~\cite{hendrycks2019augmix} to apply training-time chained augmentation mixing. 

\subsection{Results and Discussion}

We evaluate the quality and effectiveness of our generated dataset with the EndoVis-2018~\cite{allan20202018} and EndoVis-2017~\cite{allan20192017} datasets, with the latter one considered as an unseen target domain because it does not contribute to our synthetic dataset generation. The results in Table~\ref{tab:overall_res} indicate that our methods can complete the segmentation task with acceptable performance for both datasets.
As shown in Fig.~\ref{fig:qual_comp}, the instruments masks predicted by our models only have minimal visual discrepancy from the ground truth.
Considering our datasets only depend on a few trivially collected source images and get rid of gathering and annotating hundreds of real data samples, the result is promising and revolutionary for low-cost and efficient surgical instrument segmentation.

\begin{table}[!hbpt]
\centering
\caption{\textbf{Overall results of the binary surgical instrument segmentation in DSC (\%) with the EndoVis-2018 dataset and our three synthetic datasets.} AM is short for the training-time augmentation mixing. Best results of ours are shown in bold.}
\resizebox{0.8\textwidth}{!}{\begin{tabular}{c|ccc|ccc} 
\toprule
\multirow{2}{*}{Train} & \multicolumn{3}{c|}{Test on EndoVis-2018}        & \multicolumn{3}{c}{Test on EndoVis-2017}          \\ 
\cline{2-7}
                       & AM-None        & AM-Soft        & AM-Hard       & AM-None        & AM-Soft        & AM-Hard       \\ 
\hline
EndoVis-2018    & 81.58          & 83.15          & 82.91         & 83.21          & 84.06          & 83.43         \\
Synthetic-A    & 56.82          & 66.74          & 71.03          & 72.74          & 72.23          & 65.13          \\
Synthetic-B     & 57.37          & 69.42          & 72.53         & 72.65          & 73.21          & 72.41          \\
Synthetic-C     & \textbf{59.28} & \textbf{71.48} & \textbf{73.51}  & \textbf{74.37} & \textbf{75.69} & \textbf{75.16}   \\
\bottomrule
\end{tabular}
}
\label{tab:overall_res}
\end{table}

\subsection{Ablation Studies}

To show the efficacy of our training-time chained augmentation mixing, we first conduct experiments with a relevant data augmentation technique - ColorJitter, which randomly changes the brightness, contrast, and saturation of an image. Training with the Synthetic-C dataset, our augmentation strategy outperforms ColorJitter significantly with 5.33\% and 4.29\% of DSC gain on EndoVis-2018~\cite{allan20202018} and EndoVis-2017~\cite{allan20192017} datasets.

\begin{figure}[!hbpt]
\centering
\includegraphics[width=0.8\textwidth]{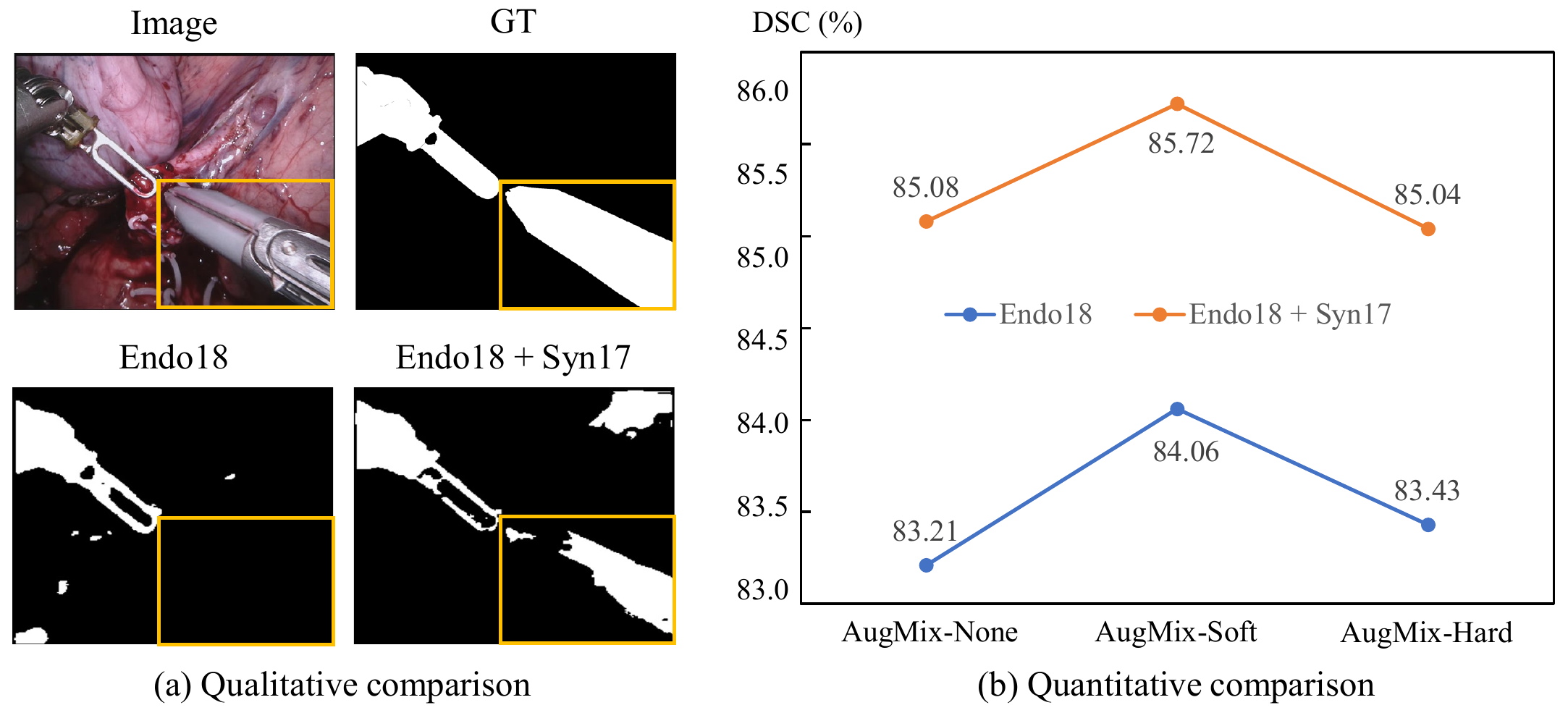}
\caption{\textbf{Qualitative and quantitative results of the class-incremental case.} The novel instrument Vessel Sealer is highlighted with the yellow rectangle in (a) . The overall performance on EndoVis-2017~\cite{allan20192017} gets greatly improved as shown in (b).}
\centering
\label{fig:clsinc}
\end{figure}

We then study the effectiveness of training with synthetic data in handling the class-incremental issue in the deployment domain. Compared with EndoVis-2018~\cite{allan20202018} dataset, there are two novel instruments in EndoVis-2017~\cite{allan20192017}, namely the Vessel Sealer and the Grasping Retractor. Following our proposed framework in Fig.~\ref{fig:blend}, we generate 2000 synthetic images for the novel instruments and combine them with EndoVis-2018~\cite{allan20202018} for training. As indicated in the highlighted area of Fig.~\ref{fig:clsinc}(a), the model manages to handle the class-incremental problem to recognize the Vessel Sealer, with only minimal effort of adding synthesized images. The overall performance on the test domain improves significantly, as shown in Fig.~\ref{fig:clsinc}(b). 

While sufficient well-annotated datasets are not common in practice, a few high-quality data samples are normally feasible to acquire. We further investigate the effect of introducing a small portion of real images when training with synthetic data. We randomly fetch 10\% and 20\% of the EndoVis-2018~\cite{allan20202018} dataset and combine it with our Synthetic-C dataset. The results in Table~\ref{tab:synreal} indicate that only a small amount of real data could provide significant benefits. Compared with training with the real EndoVis-2018~\cite{allan20202018} dataset, the models from the synthetic-real joint training scheme can efficiently achieve similar performance regarding adaptation and generalization.

\begin{table}[!hbpt]
\centering
\caption{\textbf{Results of synthetic-real joint training.} Adding only a small portion of real data can greatly improve the segmentation performance.}
\resizebox{0.7\textwidth}{!}{\begin{tabular}{c|cccc}
\hline
\multirow{3}{*}{Test} & \multicolumn{4}{c}{Train}                                                                                                                                                                                                       \\ \cline{2-5} 
                      & \multicolumn{1}{c|}{Synthetic-C} & \multicolumn{1}{c|}{\begin{tabular}[c]{@{}c@{}}Synthetic-C\\ + 10\% Endo18\end{tabular}} & \multicolumn{1}{c|}{\begin{tabular}[c]{@{}c@{}}Synthetic-C\\ + 20\% Endo18\end{tabular}} & Endo18 \\ \hline
EndoVis-2018                  & \multicolumn{1}{c}{73.51}       & \multicolumn{1}{c}{80.65}                                                               & \multicolumn{1}{c}{82.45}                                                               & 82.91  \\ 
EndoVis-2017                  & \multicolumn{1}{c}{75.16}       & \multicolumn{1}{c}{82.10}                                                                & \multicolumn{1}{c}{82.48}                                                               & 83.43  \\ 
\hline
\end{tabular}}
\label{tab:synreal}
\end{table}

\section{Conclusion}
In this work, we reevaluate the surgical instrument segmentation and propose a cost-effective data-centric framework for synthetic dataset generation. 
Extensive experiments on two commonly seen real datasets demonstrate that our high-quality synthetic datasets are capable of surgical instrument segmentation with acceptable performance and generalization ability. 
Besides, we show that our method can handle domain shift and class incremental problems and greatly improve the performance when only a small amount of real data is available. 
Future work may be extended to more complicated instrument-wise segmentation and other medical applications.
Besides, by considering more prior knowledge in practical surgical scenes, such as cautery smoke and instruments shadow, the quality of the synthetic dataset can be further improved. 

\subsubsection{Acknowledgements.}
This work was supported by the Shun Hing Institute of Advanced Engineering (SHIAE project BME-p1-21) at the Chinese University of Hong Kong (CUHK), Hong Kong Research Grants Council (RGC) Collaborative Research Fund (CRF C4026-21GF and CRF C4063-18G), (GRS)\#3110167 and Shenzhen-Hong Kong-Macau Technology Research Programme (Type C 202108233000303).

\bibliographystyle{splncs04}
\bibliography{references}

\end{document}